\documentclass[sigconf]{acmart}
\usepackage{subfigure}
\usepackage{algorithm}
\usepackage{algorithmic}
\usepackage{xcolor}
\usepackage{enumitem}

\newcommand{\method}{\sf SegTran}

\AtBeginDocument{%
  \providecommand\BibTeX{{%
    \normalfont B\kern-0.5em{\scshape i\kern-0.25em b}\kern-0.8em\TeX}}}

\copyrightyear{2020}
\acmYear{2020}
\setcopyright{acmcopyright}
\acmConference[CIKM '20]{Proceedings of the 29th ACM International Conference on Information and Knowledge Management}{October 19--23, 2020}{Virtual Event, Ireland}
\acmBooktitle{Proceedings of the 29th ACM International Conference on Information and Knowledge Management (CIKM '20), October 19--23, 2020, Virtual Event, Ireland}
\acmPrice{15.00}
\acmDOI{10.1145/3340531.3411977}
\acmISBN{978-1-4503-6859-9/20/10}

\begin{document}
\fancyhead{}
\title{Semi-Supervised Graph-to-Graph Translation}




\author{Tianxiang Zhao, Xianfeng Tang, Xiang Zhang, Suhang Wang}
\email{{tkz5084, xut10, xzz89, szw494}@psu.edu}
\affiliation{%
   \institution{College of Information Science and Technology, Penn State University}
   \city{State College}
   \country{The USA}
}

\begin{abstract}
 Graph translation is very promising research direction and has a wide range of potential real-world applications. Graph is a natural structure for representing relationship and interactions, and its translation can encode the intrinsic semantic changes of relationships in different scenarios. However, despite its seemingly wide possibilities, usage of graph translation so far is still quite limited. One important reason is the lack of high-quality paired dataset. For example, we can easily build graphs representing peoples' shared music tastes and those representing co-purchase behavior, but a well paired dataset is much more expensive to obtain. Therefore, in this work, we seek to provide a graph translation model in the semi-supervised scenario. This task is non-trivial, because graph translation involves changing the semantics in the form of link topology and node attributes, which is difficult to capture due to the combinatory nature and inter-dependencies. Furthermore, due to the high order of freedom in graph's composition, it is difficult to assure the generalization ability of trained models. These difficulties impose a tighter requirement for the exploitation of unpaired samples. Addressing them, we propose to construct a dual representation space, where transformation is performed explicitly to model the semantic transitions. Special encoder/decoder structures are designed, and auxiliary mutual information loss is also adopted to enforce the alignment of unpaired/paired examples. We evaluate the proposed method in three different datasets.

\end{abstract}

\maketitle

\section{Introduction}
\begin{figure}[t!]
  \centering
    
    \includegraphics[width=0.42\textwidth]{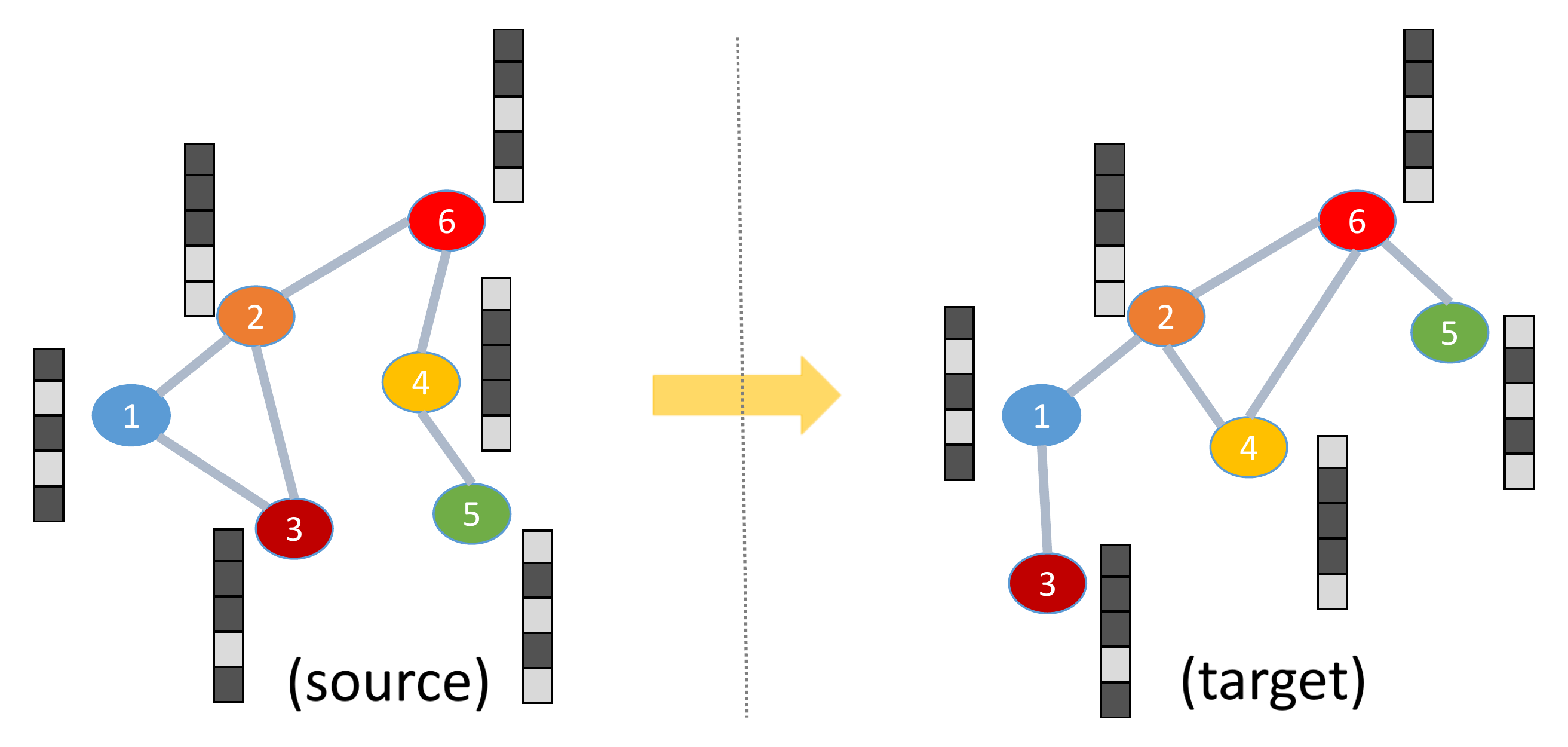}
    \vskip -1em
    \caption{An example of graph-to-graph translation} \label{fig:example}
    \vskip -1em
\end{figure}

Graph-to-Graph translation aims at transforming a graph in the source domain to a new graph in the target domain, where different domains correspond to different states. 
Figure~\ref{fig:example} gives an illustration of the graph-to-graph translation process. The graph in the source domain depicts shared music tastes of users, with each attributed node representing a user and the user's portraits. We want to translate the attributed graph of shared music tastes to co-watch graph in the target domain, which represents similar reading preference. The translation function is expected to generate the target graph based on the source graph. It is in effect learning to capture the intrinsic attributes, like aesthetic preferences and personal characters in this case, and discovering the patterns behind that domain transition. It can facilitate many real-world applications. For instance, given the brain network of a healthy subject, we may want to predict the brain network when the subject has certain disease. Given the traffic flow of a city, we want to estimate the traffic flow when events such as concert is held in the city.

The popularity of graph translation is attracting increasing attention and several efforts have been made. Most of them focus on modeling the transition of graph topology in the form of edge existence and node attributes across the source and the target domains.
For example, ~\cite{Guo2018DeepGT} formally defined this task and proposed a basic framework based on edge-node interactions. ~\cite{Guo2019DeepMG} extended previous work by building two parallel paths for updating nodes and edges respectively, and introduced a graph frequency regularization term to maintain the inherent consistency of predicted graphs. There are also some other works focusing on special needs of different applications. For example, ~\cite{Do2018GraphTP} introduced domain knowledge into the translator design for reaction prediction, and ~\cite{Shi2020AGT} adopted multi-head attention to capture the patterns in skeleton mapping.  

Despite their dedications, all aforementioned graph-to-graph translation algorithms investigate fully-supervised setting, i.e., large-scale paired graphs in source and target domain are available to provide supervision for learning translation patterns. 
However, for many domains, obtaining large-scale paired graphs is a non-trivial task, sometimes even impossible. The source and target graphs are expected to share the exactly same node sets, which is expensive to guarantee and collect in large scale. For example, in Figure~\ref{fig:example}, the source-domain graphs can be obtained from music streaming platforms like Spotify, and the target domain from e-book platforms like Amazon Kindle. It is difficult to build the corresponding pairs, as users could use different IDs across those two platforms. Another example is the brain networks, where two domains are brain activity patterns of healthy people and Dyslexia patients respectively. In this case, constructing pairs would need the same person in both healthy and diseased situation, which is impossible.

Although large-scale paired graph dataset is difficult to collect, dataset with limited number of paired graphs and large number of unpaired graphs is much easier to build, which enables us to explore semi-supervision techniques addressing this problem. Graph translation follows the classical auto-encoder framework, with an encoder in source domain and a decoder in target domain. Through introducing auxiliary reconstruction tasks and building a source-domain decoder and a target-domain encoder, those mono-domain graphs can be utilized to boost the learning process. Recovering the original source graph imposes a higher requirement on the source-domain encoder to extract more complete and representative features, and reconstructing the graph based on the its embedding can also guide the target-domain decoder to model the real graph distributions in the that domain. However, directly extending previous works in this manner has its pitfalls. The source and target domains are assumed to share one same embedding space, and the extracted embedding of source graphs are used for both reconstruction task and translation task, which makes it difficult to model the semantic drifts across two domains. Take \ref{fig:example} for example, embedding of node 2 would be close to node 3 as a result of message-passing mechanism in the source domain, but they are supposed to be distant in order to recover the non-connection information in the target domain. This trade-off could impair model's expressive ability and lead it to get sub-optimal performance. 

To cope with this semantic gap, we design a new framework, which we call as {\method}, as shown in Figure 2.  Concretely, we design a specific translation module to model the semantic transition explicitly and enforce the alignment across two domains. This translation module is trained on the limited number of paired graphs, while other components can benefit from both paired cross-domain graphs and unpaired mono-domain graphs during training. Furthermore, to assure that patterns captured by the translation module is general, we also explored providing auxiliary training signals for it by maximizing the mutual information between the source graph and its translation results. The structures of the encoder and decoder are also specially designed, as graph translation facing some intrinsic difficulties, due to diverse topology and the inter-dependency between nodes and edges~\cite{Guo2018DeepGT}. The main contributions of the paper are: 

\begin{itemize}
    \item We propose a new framework for graph translation which is better at coping with the semi-supervised scenario. It can promote future research of graph translation algorithms as the lack of large scale paired dataset is one key obstacle its exploration and applications.
    \item We design novel encoder/decoder architectures by using position embedding, multi-head attention, along with an explicit translation module, to achieve higher expressive ability. The design of each component is well-justified.
    \item Experiments are performed on three large datasets, and our model achieved the state of art result on all of them. Extensive analysis of our model's behavior is also presented.
\end{itemize}

The rest of the paper are organized as follows. In Sec.~\ref{sec:related_work}, we review related work. In Sec.~\ref{sec:problem_definition}, we formally define the problem. In Sec.~\ref{sec:methodology}, we give the details of {\method}. In Sec.~\ref{sec:experiments}, we conduct experiments to evaluate the effectiveness of {\method}. In Sec.~\ref{sec:conclusion}, we conclude with future work.

\section{Related Work} \label{sec:related_work}
In this section, we review related work, which includes graph translation and semi-supervised translation.

\subsection{Graph Translation}

Translation of complex structures has long been a hot research topic. Sequence-to-sequence learning is a standard methodology for many applications in natural language processing domain~\cite{Sutskever2014SequenceTS,Bahdanau2015NeuralMT}. Tree-structured translation was found to be beneficial for molecular generation tasks in the chemistry field~\cite{Jin2018JunctionTV,Jin2019LearningMG}. With the development of novel architectures, especially transformer~\cite{vaswani2017attention} and graph convolution network~\cite{kipf2016semi}, the power of deep network in modeling inter-dependency and interactions have been greatly improved, which make it possible to translate a more general and flexible structure, the graph. 

In graph, information are contained in both nodes and edges, and its translation has some intrinsic difficulties. First is the diverse topology structures. The semantics of graphs are not only encoded in each node's feature vector, but also in each edge's (non-)existence, which is combinatory in nature. Therefore, encoding a whole graph as a vector is often found to be brittle~\cite{Guo2018DeepGT}. Second, nodes and edges are inter-dependent, and they interact in an iterative way~\cite{You2018GraphRNNGR}. This 'recurrent' relationship is difficult to model, because it could easily go 'explode' or 'shrink', and make the learning process unstable. Third, the relationship between nodes and their neighbors are non-local. Say, distant nodes could still attribute to the existence of the target edge. These difficulties make it more difficult to capture the patterns and model the distributions of graphs. 

Earlier works mainly focus on obtaining the intermediate representation during the translation process for downstream tasks, like ~\cite{Simonovsky2018GraphVAETG,Kipf2016VariationalGA}, and payed little attention to the design of special models. ~\cite{li2018multi} translates across different graph views by modeling the correlations of different node types, and ~\cite{Liang2018SymbolicGR,Do2018GraphTP} use a bunch of predefined rules to guide the target graph generation processes. All these works require domain knowledge, and their models are domain-specific. ~\cite{sun2019graph} introduces some techniques from the natural language processing domain, but their work is refrained to the topology, and is not suitable for our setting. Works more related to ours are ~\cite{Guo2018DeepGT, Guo2019DeepMG}, which both focus on building a general graph translation framework. ~\cite{Guo2018DeepGT} updates node attributes and edge existence iteratively, and adopt a GAN~\cite{Goodfellow2014GenerativeAN} to improve the translation quality. ~\cite{Guo2019DeepMG} constructs a node-edge co-evolution block, and designs a spectral-related regularization to maintain the translation consistency. 



\subsection{Semi-supervised Translation}
Semi-supervised problem, where only a small fraction of training data has supervision, is also an important research problem. Previous works have dedicated to exploit unlabeled samples in many approaches: exploiting relationship among training examples~\cite{kingma2014semi}, enhancing feature extractors with auxiliary unsupervised task~\cite{grandvalet2005semi}, reducing model bias with ensemble-based methods~\cite{laine2016temporal}, etc. However, most of these methods are proposed for discriminative models, assuming availability of a shared intermediate embedding space with limited dimensions, or expecting the output to be a vector of probability. Our task is in essence a generative task, with both the input and output lying in the same space with a very large dimension, making those approaches infeasible.

The most related tasks to our problem setting is the utilization of monolingual dataset in neural machine translation, where unpaired sentences in the target language are used to boost model's performance. The most popular method there is based on back translation~\cite{artetxe2018unsupervised,Sennrich2016EdinburghNM,Prabhumoye2018StyleTT}, where a pre-trained target-to-source translator is applied to those unpaired sentences, hence pseudo source sentences can be obtained. They train the translator by enforcing it to generate real target from those fake pseudo sentences in the source language. In this way, a cycle-consistency can be built, and it has shown to get state-of-the-art performance. However, in those tasks, the semantic across domains are the same, which enables them to use one shared embedding space for both the source and the target sentences. Say, the meaning of the sentence remains the same, only the grammar and dictionary changes. And this difference in syntax and dictionaries can be encoded in the parameters of decoder model. Similar is the case of unsupervised style transfer in computer vision field~\cite{Zhu2017UnpairedIT}, where only low-level features like texture and color can change across domains.



\section{Problem Setting} \label{sec:problem_definition}
Many real-world problems can be viewed as the graph-to-graph translation problem, i.e., given the graph of an object in source status, predict the graph of the same object in target status. For example, predict the shared reading preference of a group of people given their similarity in music taste, or generate the brain activity network of a subject after certain diseases based on it in the healthy state. 

Throughout the paper, we use $G_s^i = \{\mathcal{V}^i, \mathbf{A}_s^i, \mathbf{F}_{s}^i\}$ to denote an  attributed graph of the $i$-th object in source status, where $\mathcal{V}^i=\{v_1^i,\dots,v_{n_i}^i\}$ is a set of $n_i$ nodes, $\mathbf{A}_{s}^i \in \mathbb{R}^{n_i \times n_i}$ is the adjacency matrix of $G_s^i$, and $\mathbf{F}_{s}^i \in \mathbb{R}^{n_i \times d_s}$ denotes the node attribute matrix, where $\mathbf{F}_{s}^i(j) \in \mathbb{R}^{1 \times d_s}$ is the node attributes of node j and $D$ is the dimension of the node attributes. Similarly, we define the corresponding target graph of $i$ as $G_t^i = \{\mathcal{V}^i, \mathbf{A}_{t}^i, \mathbf{F}_{t}^i\}$, where $\mathbf{A}_{t}^i \in \mathbb{R}^{n_i \times n_i}$ is the adjacency matrix of $G_t^i$, and $\mathbf{F}_{t}^i \in \mathbb{R}^{n_i \times d_t}$ with $d_t$ being dimensionality of node attribute in target status. We set $\mathbf{F}_{t}^i$ to be the same as $\mathbf{F}_{s}^i$ in this work due to the dataset, but our model can also be used to the scenarios when they are different. Note that we assume $G_s^i$ and $G_t^i$ share the same set of nodes while their graph structures are different. This is a reasonable assumption for many applications. Still take Figure 1 for an example, the nodes, referring to users, have to remain in the same set to build the correspondence across two domains. But the edges, representing uses' relationship with each other, along with node attributes can be different. For two graphs of different objects in the same status, say $G_s^i$ and $G_s^j$, we consider $\mathcal{V}_s^i \ne \mathcal{V}_s^j$ because $\mathcal{V}_s^i = \mathcal{V}_s^j$ is a special case, which can also be handled by our model.

To learn a model that can predict graph-to-graph translation, we need a set of paired graphs to provide supervision. We use $\mathcal{G}^p = \{G_s^i, G_t^i\}_{i=1}^{N_p}$ to denote $N_p$ paired graphs. However, obtaining large-scale paired graphs is not easy. In many cases, it is difficult or even impossible to acquire the representation of one graph in two domains, like the reading/music preference example and the brain network example. Thus, $N_p$ is usually small, which cannot well train a model for graph translation. Fortunately, it is relatively easy to obtain a set of unpaired graphs, i.e., graph in one status with its corresponding graph in another status missing. For simplicity, we use $\mathcal{G}^s = \{G_{s}^i\}_{i=1}^{N_{s}}$ to denote a set of $N_s$ graphs in the $s$ status, and $\mathcal{G}^t = \{G_{t}^i\}_{i=1}^{N_{t}}$ to represent a set of $N_t$ graphs in the $t$ status. $\mathcal{G}^p = \{G_s^i, G_t^i\}_{i=1}^{N_p}$ is the overlapping part of $\mathcal{G}^s$ and $\mathcal{G}^t$ with $N_p < N_s$ and $N_p < N_t$. 
With the above notations, the problem of semi-supervised graph translation is formally defined as
\vskip 0.5em

\noindent{}\textit{Given $\mathcal{G}^s$, $\mathcal{G}^t$ and $\mathcal{G}^p = \{G_s^i, G_t^i\}_{i=1}^{N_p}$, we aim to learn a function $f$ that can transform a source-domain graph to the target domain, i.e.,
\begin{equation}
    f(G_s^{i}) \rightarrow G_t^{i}
\end{equation}
}

\vskip -0.5em
Note that though we only consider two statuses, it is straightforward to extend our setting to multi-status graph translation.

\begin{figure*}[t!]
  \centering
    \includegraphics[width=0.9\textwidth]{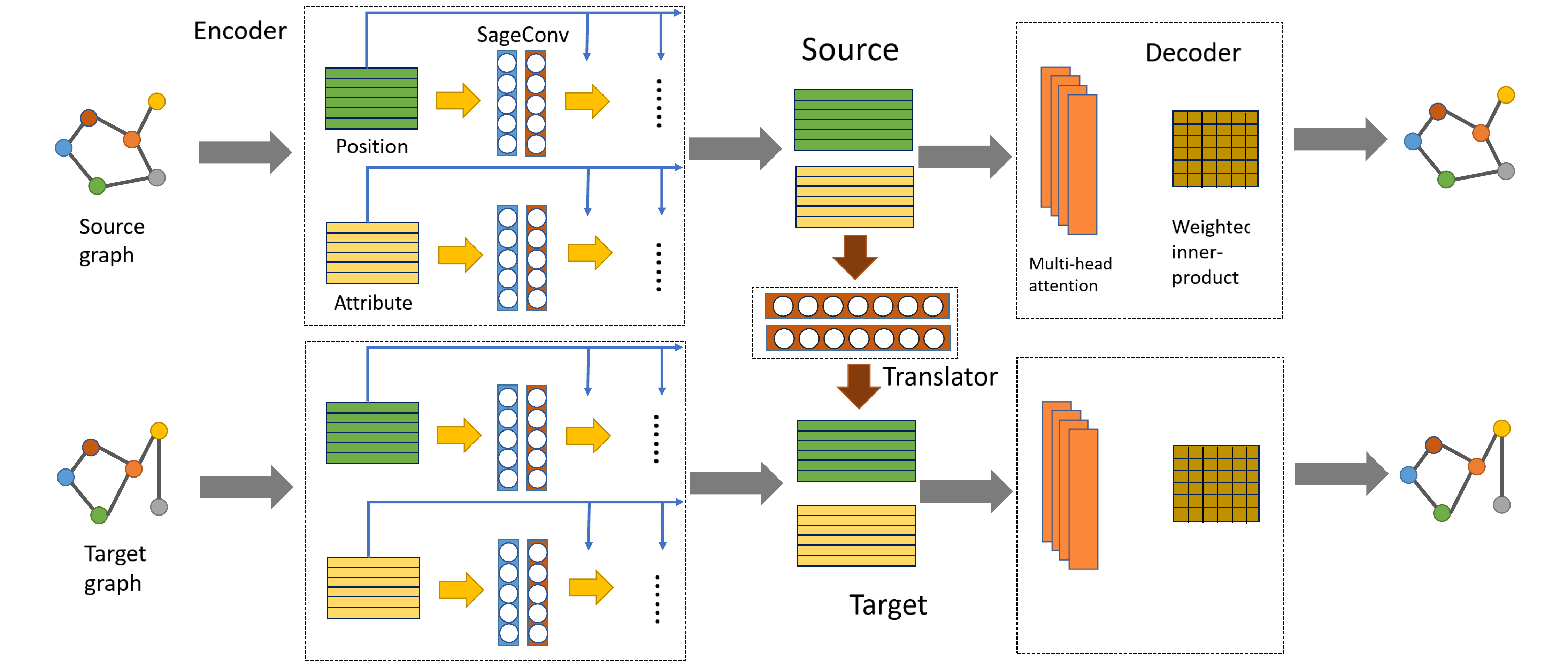}
    \vskip -1em
    \caption{Overview of the framework} \label{fig:model_architecture}
  \setlength{\abovecaptionskip}{0cm}
\end{figure*}
\section{Methodology} \label{sec:methodology}
In this section, we give the details of the proposed framework {\method} for semi-supervised graph-to-graph translation. 
It is challenging to directly translate a graph from one domain to another domain because (i) the graph structures are complex and discrete; and (ii) we have limited paired graphs to provide supervision. Thus, our basic idea is to learn good representations that preserves the topological and nodal features, translate the graph in latent domain and project the translated representation back to graph in target domain. 

An illustration of the proposed framework is shown in Figure~\ref{fig:model_architecture}. It is composed of three components: (i) two encoders which aim to learn node and graph representations for graphs of each domain; (ii) two decoders which aim to reconstruct the attributed graph to guide the encoder to learn good representations during training phase and to predict the graph in target domain during test phase; and (iii) a translator which is designed to translate the graph in the latent domain from both node and graph level. In addition, the translator leverages both the paired and unpaired graph for learning better transition ability. This design has several advantages: (i) It allows the unpaired graphs to provide auxiliary training signals for learning better encoder; (ii) Through sharing the same decoder for both translation and reconstruction tasks, its ability in modeling the graph distribution of the target domain can also be enhanced; and (iii) Due to the semantic gap issue, instead of constructing one shared embedding space and expect it to be domain-agnostic, we explicitly modeling the semantic transition process through a translator module, and form the so-called ``dual embedding'' between the source and target domain. Next, we give the details of each component.


Given an input attributed graph, we first introduce an encoder to learn node embedding that capture the network topological and node attributes so that we can conduct translation in the latent space. Graph neural networks (GNNs) have shown promising results for learning node representations \cite{kipf2016semi,Hamilton2017InductiveRL,jin2020graph,tang2020graph,jin2020self}. Thus, we choose GNN as the basic encoder. The basic idea of GNN is to aggregate a node's neighborhood information in each layer to update the representation of the node, which can be mathematically written in a message-passing framework~\cite{gilmer2017neural}:
\begin{equation}
    \mathbf{h}_v^l = \sigma(Linear(\mathbf{h}_v^{l-1}, MEAN(\mathbf{h}_w^{l-1},w \in \mathcal{N}(v)) )).
\end{equation}

$\mathbf{h}_v^l$ is the embedding of node $v$ at layer $l$, $w$ belongs to $v$'s neighbor groups $\mathcal{N}(v)$, and $\sigma$ refers to the activation function. We use a linear layer as the updating function and mean as message function respectively.
However, previous works have found two main issues of classical GNNs. The first is that their expressiveness is significantly constrained by the depth. For example, a k-layer graph convolution network(GCN) can only capture k-order graph moments without the bias term~\cite{dehmamy2019understanding}, and each layer is effectively a low-pass filter and resulting in too smooth graph features as it goes deep~\cite{wu2019simplifying}. This issue would weaken GNN's representation ability, making it harder to encode topology information in the processed node embedding. The second issue is that most of them are position-agnostic. Classical GNNs fail to distinguish two nodes with similar neighborhood but at different positions in the graph~\cite{you2019position,zhao2019hashing}. This problem becomes severer when we are using GNN to extract node embeddings to predict the existence of edges. For example, if both nodes $v, v'$ and their neighborhoods $\mathcal{N}(v), \mathcal{N}(v')$ have similar attributes, the encoder would tend to produce similar embedding for them. This makes it hard for the decoder to learn that $v$ is not connected to $N(v')$ but $v'$ is. 

\begin{figure}[t!]
  \centering
    \includegraphics[width=0.42\textwidth]{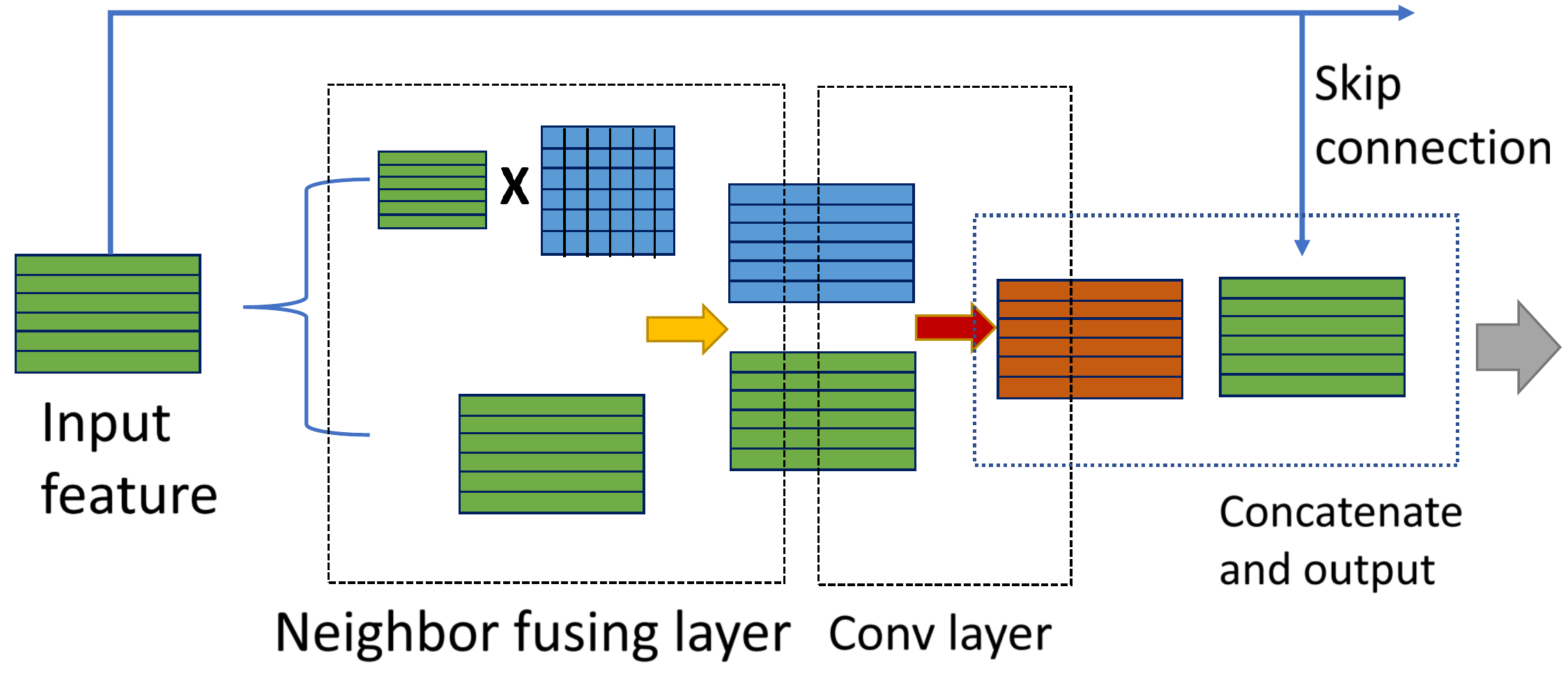}
    \vskip -1em
    \caption{Illustration of the encoder} \label{fig:encoder}
  \setlength{\abovecaptionskip}{0cm}
\end{figure}

To address these issues, we extend classical message-passing based GNNs by adding skip connections and using position embedding, as shown in Figure~\ref{fig:encoder}. With skip connections, higher GNN layers can still access lower-level features, and better representations can be learned. Concretely, we concatenate the initial node attributes to the output of each encoding layer, as in Figure~\ref{fig:encoder}. As to the position embedding, inspired by the work of ~\cite{you2019position}, we represent each node using the lengths of their shortest paths to a couple of selected anchors. Concretely, each time a graph is to be inputted, say $G_s^i$, we randomly selected $k$ nodes as anchors, and calculate position embedding of each node basing on them. We use $\mathbf{Pos}_s^i \in \mathbb{R}^{n_i \times k}$ to denote the initial position embedding of it. As proven by their work, bound of position distortions can be calculated for an anchor set of the size $O(log^{2}N)$, where $N$ is the number of nodes in that graph. Considering the size of graphs in our dataset, we preset it to eight for simplicity. Although the absolute values of obtained position embedding would be different when the anchor set changes, their relative relationship remains the same, and can be exploited. 

\subsection{Encoder}

With these preparations, now we can present the formulations of our extended GNN structure. In $l$-th block, the message passing and fusing process can be written as:
\begin{equation}
\mathbf{h}_{v}^{l}  = CONCAT(\sigma(\mathbf{W}_{F}^l \cdot CONCAT(\mathbf{h}_{v}^{l-1}, \mathbf{H}^{l-1} \cdot \mathbf{A}[:,v])), \mathbf{F}_{v}),
\end{equation}
\begin{equation}
\mathbf{p}_{v}^{l}  = CONCAT(\sigma(\mathbf{W}_{P}^l \cdot CONCAT(\mathbf{p}_{v}^{l-1}, \mathbf{P}^{l-1} \cdot \mathbf{A}[:,v])), \mathbf{Pos}_{v}).
\end{equation}
$\mathbf{H}^{l-1} = [\mathbf{h}_1^{l-1},\mathbf{h}_2^{l-1},\cdots, \mathbf{h}_{n}^{l-1}]$ represents node embedding at layer $l-1$, $\mathbf{A}[:,v]$ is the $v$-th column in adjacency matrix, and $\mathbf{P}^{l-1} = [\mathbf{p}_1^{l-1},\mathbf{p}_2^{l-1},\cdots, \mathbf{p}_{n}^{l-1}]$ is the position embedding at that layer. $\mathbf{F}$ and $\mathbf{Pos}$ are the initial node attributes and position embedding respectively. $\mathbf{W}_F^l$ and $\mathbf{W}_P^l$ are the weight parameters, and $\sigma$ refers to the activation function such as ReLU. 

\subsection{Decoder}
With the representation learned by the encoder, we introduce a decoder to reconstruct the graph. Note that the decoder is not only used to perform the reconstruction task during training phase, but will also be used to complete the graph translation in test phase. During testing, suppose we are given $G_s^i$, to predict $G_t^i$, it need to first go through source-domain encoder, then translator, followed by the target-domain decoder.

After the encoding part, for each node $v$, now we have two representation vectors, $\mathbf{H}_{v}$ and $\mathbf{P}_{v}$. $\mathbf{H}_{v}$ is the embedding of the node attributes, and $\mathbf{P}_{v}$ is the processed embedding for its relative position in the graph. To predict the existence of edges between two nodes, clearly both these two features are helpful. Considering that these two features carry different-level semantic meanings and are in different embedding spaces, directly concatenating them might be improper. In order to learn to fuse the position and attribute information, we construct the decoder with a number of same-structured blocks. Inside the block, we apply a multi-head attention layer~\cite{vaswani2017attention} with ``Query''/``Key'' being $\mathbf{P}$ and ``Value'' being $\mathbf{H}$. Its formulation can be written as:
\begin{equation}
\begin{aligned}
Attention(\mathbf{Q},\mathbf{K},\mathbf{V}) = softmax(\frac{\mathbf{Q}\mathbf{K}^T}{\sqrt{d_k}}) \cdot \mathbf{V} \\
head_i = Attention(\mathbf{P}\mathbf{W}_i^Q,\mathbf{P}\mathbf{W}_i^K,\mathbf{H}\mathbf{W}_i^V), \\
\mathbf{H}_O = CONCAT(head_1, head_2, \cdots, head_h) \cdot \mathbf{W}^O.
\end{aligned}
\end{equation}
$\mathbf{W}_i^Q \in \mathbb{R}^{d_P,d_k}, \mathbf{W}_i^K \in \mathbb{R}^{d_P,d_k},  \mathbf{W}_i^V \in \mathbb{R}^{d_H,d_v}$ and they are the query embedding matrix, key embedding matrix, value embedding matrix respectively. $\mathbf{W}^O$ is the output matrix which fuse the result from different heads. $d_P$/$d_H$ is the dimensionality of $\mathbf{P}$/$\mathbf{H}$, which is the obtained position/node embedding from the encoder. $d_k$ is the embedding dimension of Query/Key, and $d_v$ is the embedding dimension of Value. In this way, model can learn to utilize the position information in aggregating/updating attribute embedding in a data-driven approach.

Then, after concatenating position embedding and processed attribute embedding, we use a weighted inner production to perform link prediction:
\begin{equation}
\begin{aligned}
    \mathbf{E} &= CONCAT(\mathbf{H}_{O}, \mathbf{P}) \\
    \mathbf{A}_{pred} &= softmax(\sigma(\mathbf{E}^T \cdot \mathbf{S} \cdot \mathbf{E})).
\end{aligned}
\end{equation}
Here, $\mathbf{S}$ is the parameter matrix capturing the interaction between nodes. For node attribute prediction, we append a two-layer MLP after the multi-head attention module as:
\begin{equation}
    \mathbf{F}_{pred} = MLP(CONCAT(\mathbf{H}_O, \mathbf{P})),
\end{equation}

\begin{figure}[t!]
  \centering
    \includegraphics[width=0.42\textwidth]{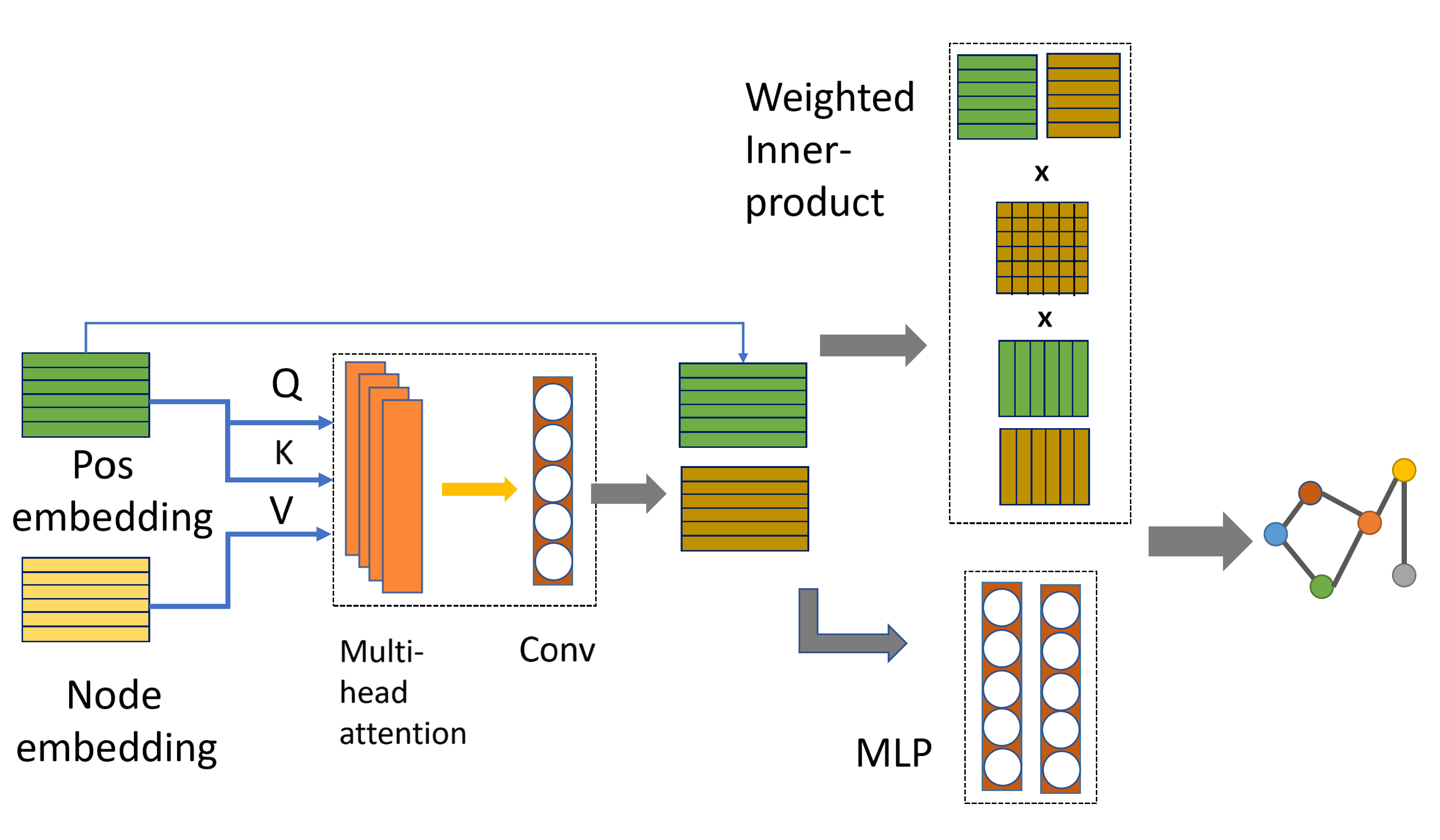}
    \vskip -1em
    \caption{Illustration of the decoder}
  \setlength{\abovecaptionskip}{0cm}
\end{figure}
Since each graph are usually sparse, i.e., the majority elements in the adjacency matrix $\mathbf{A}$ are zero, simply treating each element in the adjacency matrix equally for the graph reconstruction loss will make the loss function dominated by missing links, causing trivial results. To alleviate this problem, following existing work~\cite{pan2008one}, we assign higher weight to non-zero elements in $\mathbf{A}$ as follows:
\begin{equation}
    S_{ij} = 
    \begin{cases}
    1, & \text{if } A_{ij} > 0 \\
    \delta,              & \text{otherwise}
\end{cases}
\end{equation}
where $\delta$ is between 0 and 1 to control the weight of the missing links in a graph. With $\mathbf{S}$, the graph reconstruction loss can be written as:
\begin{equation}
\begin{aligned}
    \mathcal{L}_{rec}(G) = \|\mathbf{S} \odot (\mathbf{A}_{pred} - \mathbf{A})\|_F^2 + \|\mathbf{F}_{pred} - \mathbf{F}\|_F^2
\end{aligned}
\end{equation}
where $\odot$ denotes element-wise multiplication.



\subsection{Translator Module}
With the encoders and decoders in place, now we can introduce how we learn the translation patterns across two domains. In this framework, we model the translation process in the intermediate level, through transitioning a source graph embedding to its target graph embedding. As shown in Figure~\ref{fig:model_architecture}, Translator module is a key component in our framework, which is required to build the mapping from source domain to target domain. Concretely, we adopt a MLP structure to implement it. The translator operates in a node-wise fashion, using both global feature along with node-level feature as the input:
\begin{equation}
\begin{aligned}
  \mathbf{H}_{v,t\_pred}, \mathbf{P}_{v,t\_pred} =& MLP(CONCAT(\mathbf{H}_{v,s}, \mathbf{P}_{v,s}, \\ &READOUT(\mathbf{H}_{s}), READOUT(\mathbf{P}_{s}))).
\end{aligned}
\end{equation}
$t\_pred$ refers to the translated result, and $\mathbf{H}_{v,t\_pred}$ represent the translated intermediate-level node attribute embedding of node $v$. Same is the case for $ \mathbf{P}_{v,t\_pred}$. $READOUT$ is the global pooling function, which is used to fuse the representation of all nodes in a graph. In this way, the graph-level embedding is appended to the extracted node-level embedding, so that translation patterns can be learned with both local and global features.

During training, this correspondence is easy to be established for paired graphs. We can perform a regression task and minimize the prediction loss in the intermediate embedding level as:
\begin{equation}
    L_{trans}^{p}(G_s,G_t) = \|\mathbf{H}_{t} - \mathbf{H}_{t\_pred} \|_F^2 + \| \mathbf{P}_{t} - \mathbf{P}_{t\_pred}\|_F^2
\end{equation}
In this equation, $t$ refers to the embedding obtained from the target graph, and $t\_pred$ means the predicted result based on the source graph by the translator module. 

However, for the large amount of unpaired graphs, this kind of training signals can not be obtained. Inspired by ~\cite{belghazi2018mutual,hjelm2018learning}, we propose to use the mutual information(MI) score as an auxiliary supervision to better align these two spaces, as it can quantify the dependence of two random variables. If two graphs are paired, then they should have a high MI score with each other, and if they are not paired, then their MI score would be low. Therefore, by optimizing the MI score between the translated result and the source graph, the translator module would be encouraged to produce more credible results. However, as the dimension of embedding space is too high($\mathcal{R}^{N \cdot (d_P+d_F)}$), directly working on it could suffer from curse of dimensionality and could result in brittle results. Addressing this, we apply MI score to global-level embedding through the $READOUT$ function, which can change the dimension to $\mathcal{R}^{d_P+d_F}$. For simplicity of notation, we use $\mathbf{g}_s$ to denote the global feature of graph $G_s$ after the $READOUT$
\begin{equation}
    \mathbf{g}_s = READOUT(CONCAT(\mathbf{H}_s, \mathbf{P_s})),
\end{equation}
and use $\mathbf{g}_{t\_pred}$ to denote the output of this translator module, the global feature of translated result
\begin{equation}
    \mathbf{g}_{t\_pred} = READOUT(CONCAT(\mathbf{H}_{t\_pred}, \mathbf{P_{t\_pred}}))
\end{equation}
The mutual information score is calculated on top of that, following the same procedures as ~\cite{hjelm2018learning}, which involves a specific estimator $\mathbf{MI}$:
\begin{equation}
\begin{aligned}
    \mathcal{L}_{MI}^{u}(G_s,G_{t\_pred}) =\mathbf{MI}(\mathbf{g}_s, \mathbf{g}_{t\_pred})
\end{aligned}
\end{equation}

\subsection{Objective Function of \method}
Putting previous parts together, we can get our full model architecture. The optimization goal during training can be written as:
\begin{equation}
\begin{aligned}
  \min_{\theta} & \frac{1}{|\mathcal{G}^p|}\sum_{(G_s, G_t) \in \mathcal{G}^p} \mathcal{L}_{trans}^{p}(G_s,G_t)  \\
  & + \lambda(\frac{1}{|\mathcal{G}^s|} \sum_{G_s \in \mathcal{G}^s}\mathcal{L}_{rec}(G_s) + \frac{1}{|\mathcal{G}^t|} \sum_{G_s \in \mathcal{G}^t}\mathcal{L}_{rec}(G_t)) \\
  & +\mu \cdot \frac{1}{|\mathcal{G}^s|}(\mathcal{L}_{MI}^{u}(G_s, G_{t\_pred})).
\end{aligned}
\end{equation}
Besides the paired translation loss $\mathcal{L}_{trans}^{p}$ defined in Equation 11, we also add the reconstruction loss $\mathcal{L}_{rec}$ defined in Equation 9 and MI loss $\mathcal{L}_{MI}$ defined in Equation 14. $\mathcal{L}_{rec}$ is applied to graphs from both source and target domains, and its weight is controlled by the hyper-parameter $\lambda$. $\mathcal{L}_{MI}$ is applied to only unpaired source graphs, by computing the mutual information between $G_s$ and translation result $G_{t\_pred}$, and its weight is controlled by $\mu$.

\subsection{Training Algorithm}
Besides, as our model is composed of multiple different components, we follow a pretrain-finetune pipeline to make the training process more stable. The full training algorithm can be found in Algorithm 1. We first pre-train the encoder and decoder using the reconstruction loss with both paired and unpaired graphs, so that one meaningful intermediate embedding space for each domain can be learned. Then, we fix them, only train the translator module, to learn the mapping between two embedding spaces. After that, we fix the whole model, and prepare the mutual information estimator. When all these preparations are done, we start the fine-tune steps, by alternatively updating on paired and unpaired graphs, and train the whole model in an end-to-end manner.

\begin{algorithm}
  \caption{Full Training Algorithm}
  \label{alg:Framwork}
  \begin{algorithmic}[1] 
  \REQUIRE 
    $\{\langle G_s, G_t\rangle^p, \langle G_s \rangle^u, \langle G_t\rangle^u \}$
  \ENSURE $Reconstructed \ G_s \  and \  G_t$
    \STATE Initialize the encoder, decoder in both source and target domain, by pretraining on loss $\mathcal{L}_{rec}(G_s)$ and $\mathcal{L}_{rec}(G_t)$;
    \STATE Fix other parts, only train the translator module, based on loss $\mathcal{L}_{trans}^{p}$;
    \STATE Fix the whole model, pretrain the mutual information estimator, following ~\cite{hjelm2018learning};
    \WHILE {Not Converged}
    \STATE Receive a batch of paired graphs $\langle G_s, G_t \rangle^p$; 
    \STATE Update the model using $\mathcal{L}_{rec} + \mathcal{L}_{trans}$; 
    \STATE Update the mutual information estimator;
    \STATE Receive a batch of unpaired graphs $\langle G_s\rangle^u, \langle G_t \rangle^u$;
    \STATE Update the model using $\mathcal{L}_{rec} + \mathcal{L}_{MI}$; 
    
    \ENDWHILE
    \RETURN Trained encoder, decoder, and translator module.
  \end{algorithmic}
\end{algorithm}

\section{Experiments} \label{sec:experiments}
In this section, we conduct experiments to evaluate the effectiveness of {\method} and the factors that could affect its performance. In particular, we aim to answer the following questions.
\begin{itemize}
    \item How effective is {\method} in graph translation by leveraging paired and unpaired graphs?
    \item How different ratios of unpaired graphs could affect the translation performance of {\method}?
    \item What are the contributions of each components of {\method}?
\end{itemize}
We begin by introducing the experimental settings, datasets and baselines. We then conduct experiments to answer these questions. Finally, we analyze parameter sensitivity of {\method}. 

\subsection{Experimental Settings}
\subsubsection{Datasets} 
We conduct experiments on one synthetic dataset, BA~\cite{Guo2018DeepGT}, and two widely used real-world datasets, DBLP~\cite{Tang2008ArnetMinerEA} and Traffic~\footnote{https://data.cityofnewyork.us/Transportation/2018-Yellow-Taxi-Trip-Data/t29m-gskq}. The details of the datasets are given as follows: 
\begin{itemize}[leftmargin=*]
    \item \textbf{BA}: In BA dataset, source domain is constructed by the Barabasi-Albert model~\cite{McInnes1999EmergenceOS}. The graph is built by adding nodes to it sequentially with preferential attachment mechanism, until it reaches $40$ nodes. Each newly-added node is connected to only one existing node randomly, with probability $\frac{k_i}{\sum_j k_j}$. Here, $k_i$ means the current degree of node $i$. This method can generate graphs that follow the scale-free degree distributions. The target graph is constructed as $2$-hop reach-ability graph, i.e., if node $i$ is $2$-hop reachable from node $j$ in the source graph, then they are connected in the target graph. As the generated graphs are unattributed, we initialize the node attribute matrix $\mathbf{F}$ to be the same as adjacency matrix $\mathbf{A}$. We include this synthetic dataset to understand if {\method} can really capture the translation patterns.
    
    \item \textbf{DBLP}: DBLP is a multi-view citation network, with $22559$ nodes representing researchers. It provides edges of three types, representing co-authorship, citation, and research overlapping respectively. Each node has a $2000$-dimension vector available, encoding the research interests. In our experiment, we use the citation network as the source, and research overlapping network as the target domain. This dataset is given in the transductive learning setting, and all the nodes appeared in one single large graph. Addressing this, we manually split it by first selecting a center node and then sampling nodes within its 2-hop neighborhood to get a smaller graph. The sampled graphs are not required to have exactly same number of nodes, and we control the graph size by setting the upper-bound of node degree as $15$.
    
    \item \textbf{Traffic}: For traffic dataset, we use the publicly available New York Taxi Trip dataset in year $2015$. Each trip record contains the take-on and take-off places, as well as the trip-start time. We follow prior studies \cite{yao2018deep,yao2019revisiting,yao2019learning,wang2019simple} and split the city into $100$ regions, and build the edges based on the taxi flow amount between each region pairs within one hour. This results in $365*24$ graphs in total. On this dataset, we perform a traffic flow prediction task, and set the target domain as the graph state one hour in the future. Node attributes are initialized using the mean of historic taxi flows in the past $24$ hours.
\end{itemize}

\begin{table}[t]
  \centering
  \setlength{\tabcolsep}{4.5pt}
  \normalsize
  \caption{Statistics of Three Datasets. } \label{tab:statistics} 
  \vskip -1.5em
  \begin{tabular}{ | c | c || c  | c | c |}
    \hline
    \multicolumn{2}{|c||}{} & BA & DBLP & Traffic \\
    \hline
    \multicolumn{2}{|c||}{Dataset Size} & 5000 & 22559 & 8760 \\
    \multicolumn{2}{|c||}{Average Graph Size} & 40 & 125 & 100 \\
    \hline
    \multicolumn{2}{|c||}{Paired Training} & 500 & 2255 & 876 \\
    \multicolumn{2}{|c||}{Unpaired Training} & 2000 & 9020 & 3504 \\
    \hline
    \multicolumn{2}{|c||}{Paired Testing} & 1000 & 4510 & 1752 \\
    \hline
  \end{tabular}
\end{table}

The statistics of the three datasets are summarized in Table~\ref{tab:statistics}, where ``Dataset Size'' is the total number of graphs and ``Average Graph Size'' is the average number of nodes in each graph. Note that in BA and Traffic datasets, all graphs have the same size. As we adopt the semi-supervised problem setting, we only use a small subset of graphs as paired, and treat a larger subset as unpaired. The size of each subset is also listed in the ``Paired Training'' and ``Unpaired Training'', respectively. The number of paired graphs for testing is listed in the ``Paired Testing'' row. 

\subsubsection{Baselines}
We compare {\method} with representative and state-of-the-art supervised and semi-supervised graph-to-graph translation approaches, which includes:
\begin{itemize}[leftmargin=*]
    \item DGT~\cite{Guo2018DeepGT}: This method belongs to the encoder-decoder framework. The encoder is composed of edge-to-edge blocks to update the edge representation and edge-to-node blocks to obtain the node embedding. The decoder performs an inverse process as to the encoder, and maps extracted node representation to the target graph. It also utilizes a discriminator and follows the adversarial training approach to generate more ``real'' graphs. For implementation, we used the the code provided by the author~\footnote{https://github.com/anonymous1025/Deep-Graph-Translation-}.
    \item NEC-DGT~\cite{Guo2019DeepMG}: This is the state-of-the-art approach in graph translation, which also follows the encoder-decoder framework. To model the iterative and interactive translation process of nodes and edges, it split each of its block into two branches, one for updating the node attributes and one for updating the edge representations. Special architecture is designed for each branch. Besides, it also designed a spectral-based regularization term to learn and maintain the consistency of predicted nodes and edges. We use the implementation provided by the author~\footnote{https://github.com/xguo7/NEC-DGT}.
    \item NEC-DGT-enhanced: As both DGT adn NEC-DGT can only work on paired graphs, they are unable to learn from those unpaired training samples, which would make the comparison unfair. Addressing this problem, we design an extension of NEC-DGT by adding a source-domain decoder, and call it NEC-DGT-enhanced. With this auxiliary decoder, NEC-DGT-enhanced can utilize unpaired graphs by performing reconstruction tasks, which is supposed to improve the performance of the encoder. Note that this can be treated a a variant of {\method} without dual embedding and translator module.
    \item {\method}\_p: To further validate the performance of our model design and verify the improvement can be gained from utilizing unpaired graphs, we also designed a new baseline, {\method}\_p. It is of the same architecture as {\method}, but is trained only on the paired graphs. 
\end{itemize}



\subsubsection{Configurations}
All experiments are conducted on a $64$-bit machine with Nvidia GPU (Tesla V100, 1246MHz , 16 GB memory), and ADAM optimization algorithm is used to train all the models. 

For DGT, NEC-DGT, and NEC-DGT-Enhanced, the learning rate is initialized to $0.0001$, following the settings in their code. For {\method} and {\method}\_p, the learning rate is initialized to $0.001$. In {\method}, the value of hyper-parameter $\lambda$ and $\mu$ are both set as $1.0$. A more detailed analysis over the performance sensitivity to them can be found in Section~\ref{sec:parameter_sensitivity}. Besides, all models are trained until converging, with the maximum training epoch being $5000$. 

\subsubsection{Evaluation Metrics} For the evaluation metrics, we adopt mean squared error(MSE) and mean average percentage error(MAPE). As MSE is more sensitive when the ground-truth value is large and MAPE is more sensitive when the ground-truth value is small, we believe they can provide a more comprehensive comparison combined. Besides, class-imbalance exists in this problem, as edges are usually sparse, and directly training on them would result in trivial results. 
Addressing it, we re-weight the importance of edges and non-edges during the loss calculation, in both training and testing.

\begin{table*}[h]
  \centering
  \setlength{\tabcolsep}{4.5pt}
  \normalsize
  
  \caption{Comparison of different approaches in the semi-supervised scenario. Among these approaches, DGT, NEC-DGT and {\method}\_p use only paired graphs, while NEC-DGT-Enhanced and {\method} use both paired and unpaired graphs during training.}\label{tab:real}
  \vskip -1em
  \begin{tabular}{c || c | c || c | c || c | c }

    \hline
     &  \multicolumn{2}{|c||}{BA} &  \multicolumn{2}{|c||}{DBLP} &  \multicolumn{2}{|c}{Traffic} \\
    \hline
    Methods & MSE & MAPE & MSE & MAPE & MSE & MAPE \\
    \hline
    DGT & $0.377\pm0.0053$ & $0.281\pm0.0029$ & $0.385\pm0.0074$ & $0.276\pm0.0068$ & $0.252\pm0.0071$ & $0.183\pm0.0064$ \\
    NEC-DGT & $0.217\pm0.0042$ & $0.153\pm0.0032$ & $0.205\pm0.0026$  & $0.159\pm0.0024$ & $0.138\pm0.0274$ & $0.110\pm0.0186$ \\
    NEC-DGT-enhanced & $0.195\pm0.0045 $  & $0.131\pm0.0021$ & $0.197\pm0.0016$ & $0.148\pm0.0033$ & $0.076\pm0.0087$  & $0.065\pm0.0122$  \\
    \hline
    {\method}\_p & $0.154\pm0.0038$ & $0.119\pm0.0026$ & $0.203\pm0.0029$ & $0.151\pm0.0017$ & $0.091\pm0.0072$ & $0.067\pm0.0059$\\
    {\method}  &$\mathbf{0.132\pm0.0028}$  & $\mathbf{0.098\pm0.0011}$ & $\mathbf{0.192\pm0.0007}$ & $\mathbf{0.135\pm0.0003}$ & $\mathbf{0.067\pm0.0046}$ & $\mathbf{0.049\pm0.0027}$ \\
    \hline
  \end{tabular}
\end{table*}
\subsection{Graph-to-Graph Translation Performance}


To answer the first question, we compare the graph tranlation performance of {\method} with the baselines under the semi-supervised scenario on the three datasets. We train the model on the paired and unpaired training graphs, and conduct graph translation on the test graph. Each experiment is conducted $5$ times to alleviate the randomness. The average performance with standard deviation in terms of MSE and MAPE are reported in Table~\ref{tab:real}. From the table, we can make following observations: 
\begin{itemize}[leftmargin=*]
    \item {\method} outperforms all other approaches by a reasonable margin. Compared with NEC-DGT-Enhanced, It shows an improvement of $0.082$ on BA dataset, $0.005$ on DBLP, and $0.009$ on Traffic dataset measured by MSE loss. In the term MAPE loss, the improvements are $0.033$, $0.013$, and $0.016$ respectively. This result validates the effectiveness of {\method}.
    \item {\method} performs much more stable than NEC-DGT-Enhanced, although they both utilize unpaired graphs. Looking at the standard deviations, it can be observed that NEC-DGT-Enhanced has higher performance variations. For example, its standard deviation on BA dataset is about two times that of {\method}. This phenomena could result from NEC-DGT-enhanced's difficulty in trading-off between the semantics from source domain and the semantics from target domain during learning the intermediate embedding space. 
    \item The performance differences on DBLP dataset is smaller than those on the other two. This could result from the dataset size. Refering to Table 1, DBLP dataset is significantly larger than both other two datasets. Therefore, the bonus from adding more unpaired graphs would be smaller, which results in this result.
    \item Generally, NEC-DGT-Enhanced performs better than NEC-DGT, and {\method} performs better than {\method}\_p. This observation can test the significance of learning from unpaired graphs. NEC-DGT-Enhanced follows the exactly same structure as NEC-DGT other than an auxiliary decoder to perform reconstruction on the source-domain graphs, and it achieves an improvement of $0.022$, $0.008$ and $0.062$ respectively measured by MSE losses. Similar observations can be made in the case of comparing {\method} and {\method}\_p. This result shows that in the case when paired graphs are limited, it is important to use unpaired graphs to boost model's performance. 
\end{itemize}

To summarize, these results prove the importance of introducing unpaired graphs to graph translation tasks with only limited number of paired training samples. Besides, they also validate the intuition that explicitly modeling the difference between two domains in the intermediate space can make it easier for the model to utilize unpaired graphs.

\begin{figure}[h!]
  \centering
    \subfigure[BA]{
		\label{fig:BA}
		\includegraphics[width=0.23\textwidth]{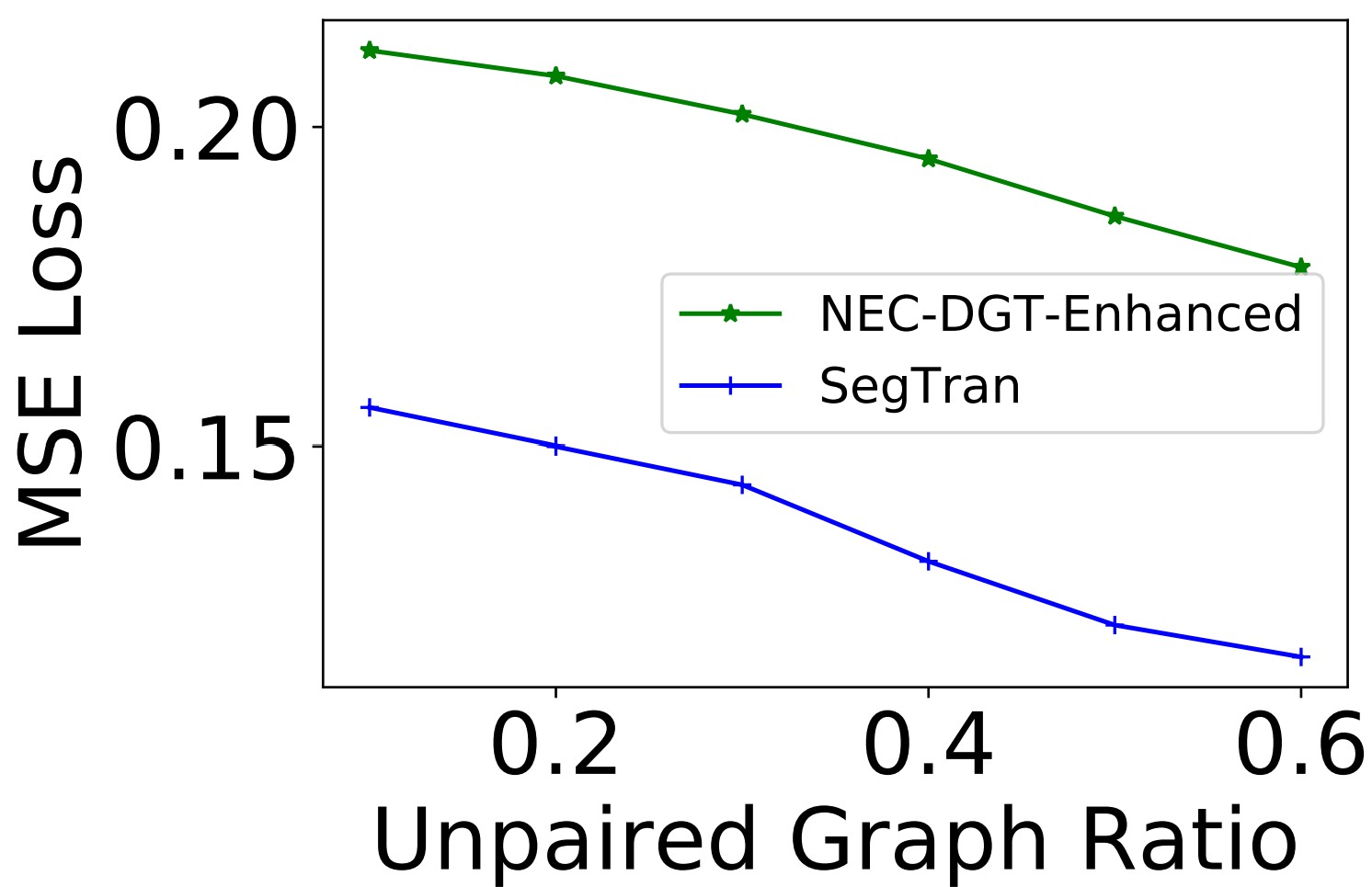}}
    \subfigure[Traffic]{
		\label{fig:traffic}
		\includegraphics[width=0.23\textwidth]{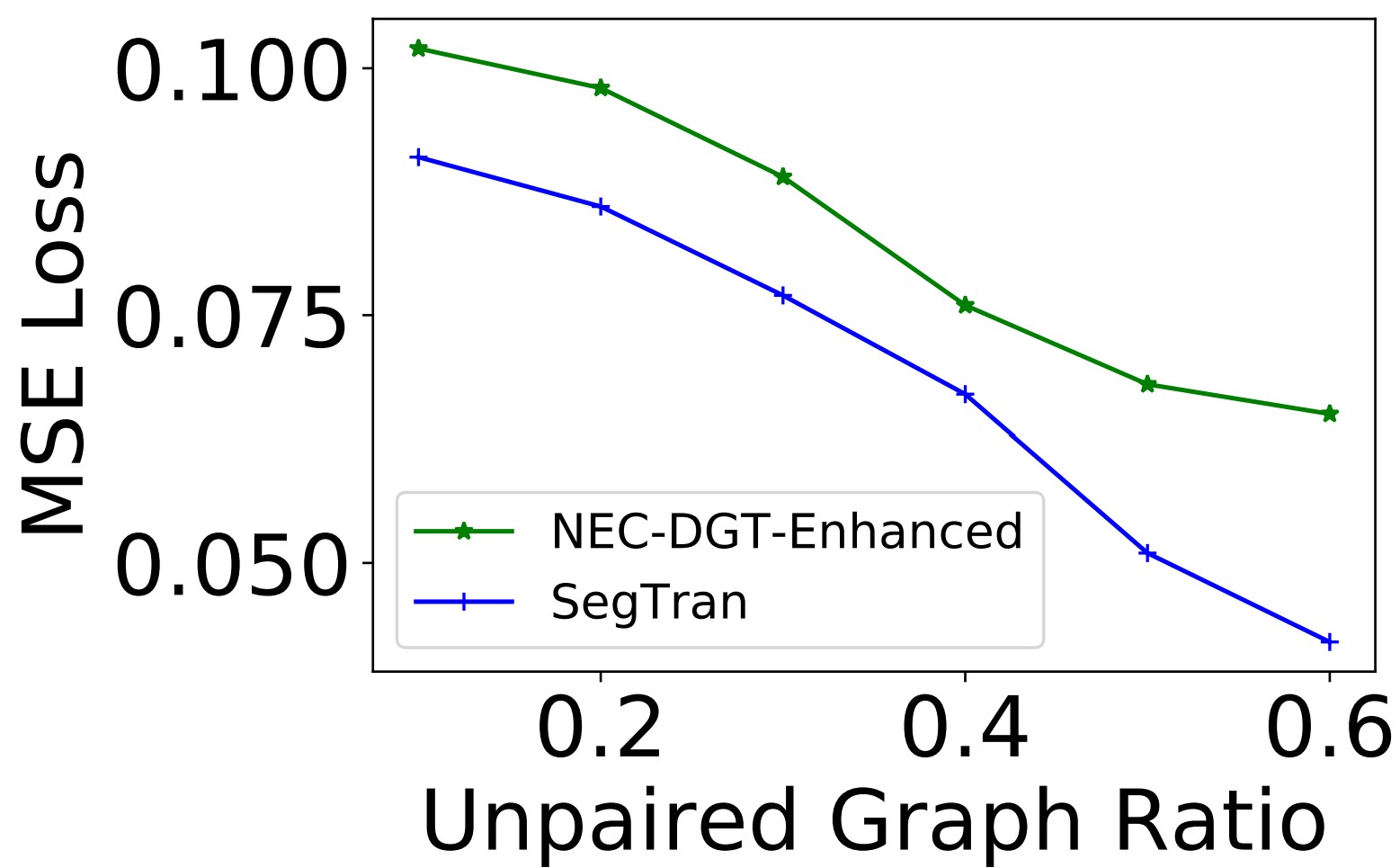}}
	\vskip -1.5em
    \caption{Affects of Ratios of Unpaired Graphs.} \label{fig:ratio_unpaired_graphs}
  \setlength{\abovecaptionskip}{0cm}
\end{figure}

\subsection{Ratio of Unpaired Graphs}
In this subsection, we analyze the sensitivity of our model and NEC-DGT-Enhanced towards the amount of unpaired graphs, which answers the second question. We fix the number of paired training graphs, and change the settings by taking different percentage of samples in the dataset as unpaired graphs for semi-supervised training. Concretely, we vary unpaired graph to paired graph ratio as $\{10\%,20\%,30\%,40\%,50\%,60\%\}$, and fix other hyper-parameters unchanged. We only report the performance on BA and Traffic as we have similar observation on DBLP. The results are presented in Figure~\ref{fig:ratio_unpaired_graphs} 
From the figure, we make the following observations:
\begin{itemize}[leftmargin=*]
    \item In general, as the increase of ratio of unpaired graphs, the performance of both NEC-DGT-Enhanced and {\method} increases, which implies that unpaired graphs can help to learn better encoder and decoder for graph-to-graph translation.
    \item As the increase of the ratio of unpaired graphs, the performance of {\method} is consistently better than NEC-DGT-Enhanced, which validate the effectiveness of {\method} in learning from unpaired graphs.
\end{itemize}


\subsection{Ablation Study}
To answer the third question, in this subsection, we conduct ablation study to understand the importance of each component of the proposed framework {\method}. All experiments are performed in semi-supervised setting, with configurations exactly same as {\method}, if not stated otherwise.


\begin{table}[t]
  \centering
  \setlength{\tabcolsep}{4.5pt}
  \normalsize
  \caption{Evaluation of the significance of different designs in our approach on BA and DBLP datasets. The scores are computed using MSE loss.} \label{tab:ablation}
  \vskip -1em
  \begin{tabular}{c || c  || c }

    \hline
    Methods & BA & DBLP\\
    & Semi sup & Semi sup \\
    \hline
    Shared Embedding & 0.196 & 0.245 \\
    No position & 0.142 & 0.204 \\
    No MI & 0.141 & 0.197  \\
    No multi-head attention & 0.135 & 0.190 \\
   {\method} & 0.132 & 0.192  \\
    \hline
  \end{tabular}
\end{table}

\textbf{Gain from dual embedding}
In our model design, ``dual embedding'' is adopted to distinguish between the source and the target domain, and help the model to learn more from unpaired graph. To test its affect, we perform an ablation study by removing the translation module, requiring two domains to share the same embedding space. Other parts are not influenced, except the mutual information loss, which is no longer needed in this case. The performance of ``Shared Embedding'' in terms of MSE is shown in Table~\ref{tab:ablation}. From the table, we can see that on both BA and DBLP dataset, compared with {\method}, removing dual embedding and changing to shared embedding would result in a significant performance drop, which is about $0.065$ and $0.053$ points decrease, respectively. This is because ``dual embedding'' can ease the trade-off between reconstruction and translation tasks. Without it, the same features would be used for them, which could result in sub-optimal performance. This result shows the effectiveness of this design in leveraging the semantic change from source domain to target domain. 

\textbf{Importance of position embedding}
In this part, we test the effects of the position embedding, and evaluate its contributions in guiding the encoding process. In removing the position embedding, we leave the model architecture untouched, and use the first eight dimensions of node attributes to replace the calculated position embedding. Through the result in Table~\ref{tab:ablation}, it can be seen that it will result in a performance drop of $0.01$ on BA, and $0.012$ on DBLP. This result shows that position embedding is important for extending representation ability of graph encoder module.

\textbf{Importance of MI loss}
We also test the benefits from auxiliary mutual information loss when aligning the transformed embedding of unpaired graphs. For this experiment, we simply remove this loss during the fine-tune steps, and observe the change of performance. We find a drop of $0.009$ points on BA, and $0.005$ on DBLP. The drop is smaller on DBLP datast, and it may still be a result of dataset size. As DBLP is much larger, the number of paired graphs is more sufficient for the training, which makes this auxiliary loss not that important. But overall, we can observe that this supervision can train the translator better.

\textbf{Influence of multi-head attention}
In this part, we test the influence of the multi-head attention module in the decoding process. This module is designed to fuse the processed embedding of nodes based on their relative distances. The relative distances is measured using the processed position embedding along with the metric space constructed by each attention head. Through multi-head attention, higher-order interaction between processed node embedding and position embedding is supported. For comparison, we directly remove this layer from the decoder. This manipulation will not influence other parts of the translation network. From Table~\ref{tab:ablation}, we can see that on BA dataset, this module can bring a moderate improvement of around $0.003$. On DBLP dataset, the performance is similar whether this module is removed or not. Therefore, the importance of this layer is rather dependent on the complexity of dataset, and sometimes it is not necessary. 

\begin{figure}[h]
  \centering
    \subfigure[MSE]{
		\label{fig:BA}
		\includegraphics[width=0.22\textwidth]{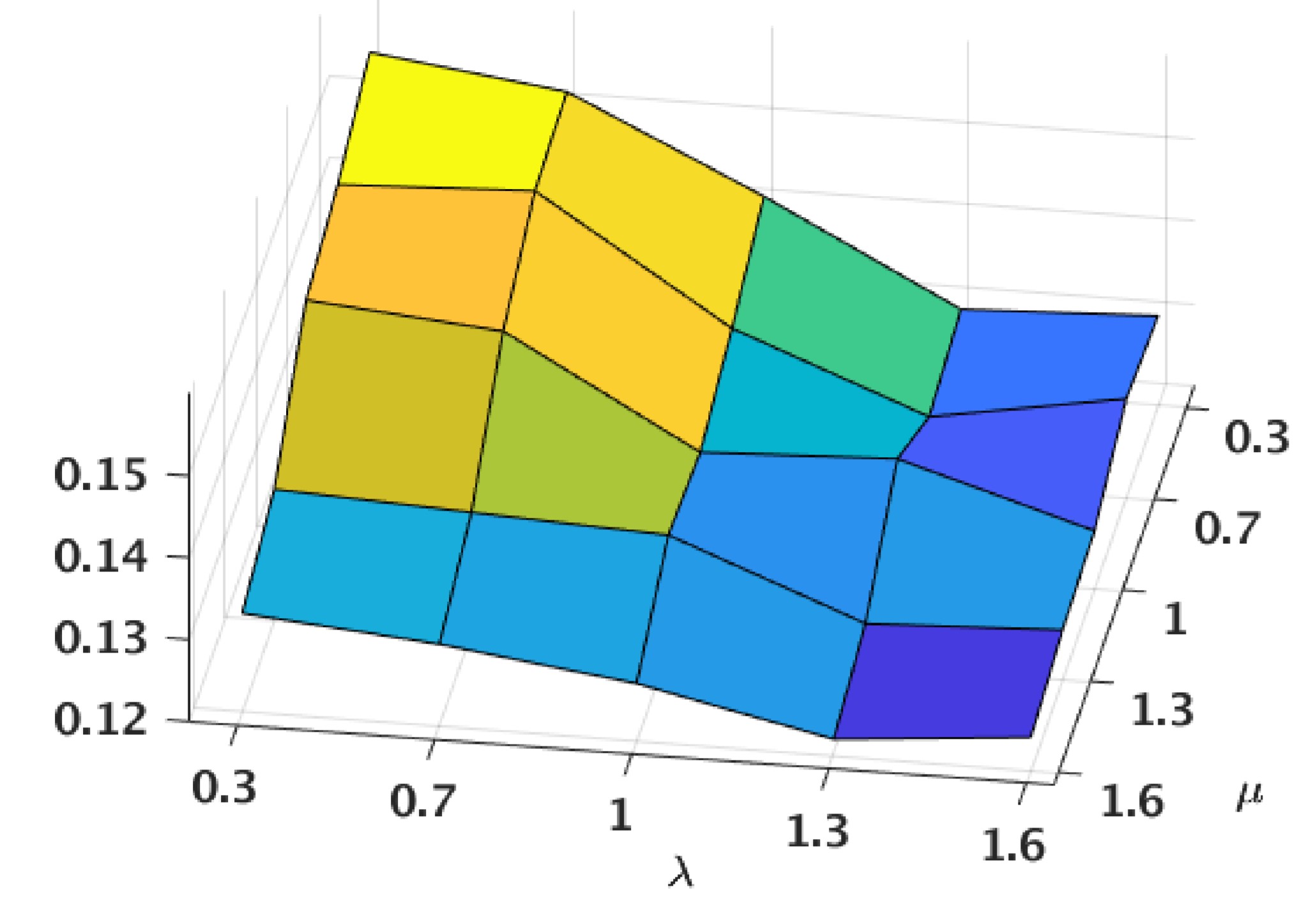}}
    \subfigure[MAPE]{
		\label{fig:traffic}
		\includegraphics[width=0.22\textwidth]{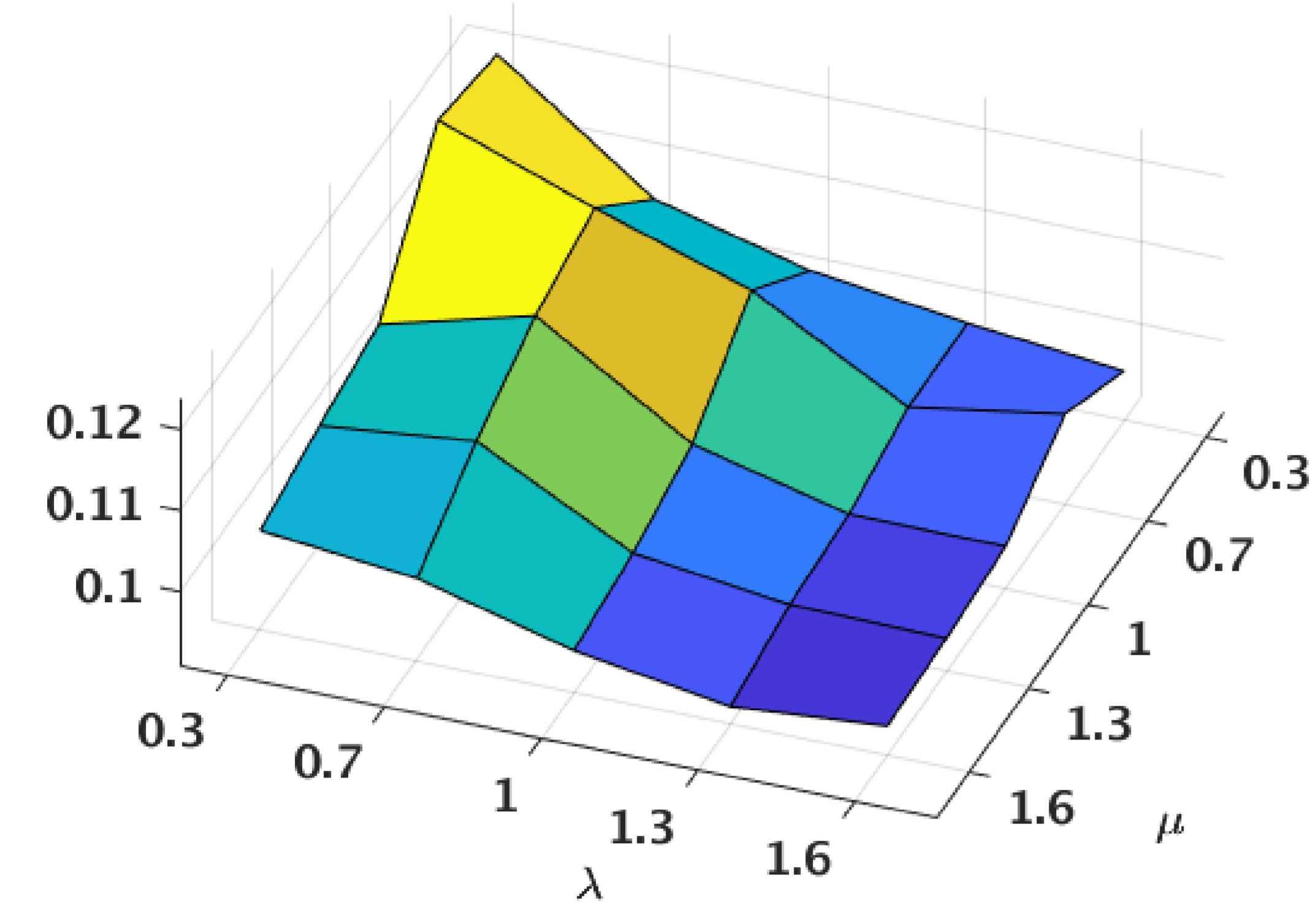}}
	\vskip -1em
    \caption{Parameter Sensitivity on BA} \label{fig:parameter_sensitivity}
\end{figure}

\subsection{Parameter Sensitivity Analysis} \label{sec:parameter_sensitivity}
In this subsection, we analyze the sensitivity of {\method}'s performance towards the hyper-parameters $\lambda$ and $\mu$, the weight of reconstruction loss and mutual information loss. We vary both $\lambda$ and $\mu$ as $\{0.3,0.7,1.0,1.3\}$. These experiments all use the same pre-trained model, as these two hyper-parameters only influence fine-tune process. Other settings are the same as {\method}. This experiment is performed on BA dataset, and the result is shown in Figure~\ref{fig:parameter_sensitivity}. The $z$ axis is the translation error measured using MSE loss, $x$ axis refers to the value of $\lambda$, and $y$ axis represents $\mu$.


From the figure, we can observe that generally, reconstruction loss and mutual information loss are both important for achieving a better performance. When $\lambda$ is small, it is difficult for this model to achieve a high performance. This observation makes sense because reconstruction loss guides the encoder to extract more complete features from input graphs, and consequently has a direct influence on the quality of intermediate embedding space. 


\subsection{Case study}
\begin{figure}[h!]
  \centering
  \vskip -1em
    \subfigure[Source]{
    \includegraphics[width=0.18\textwidth]{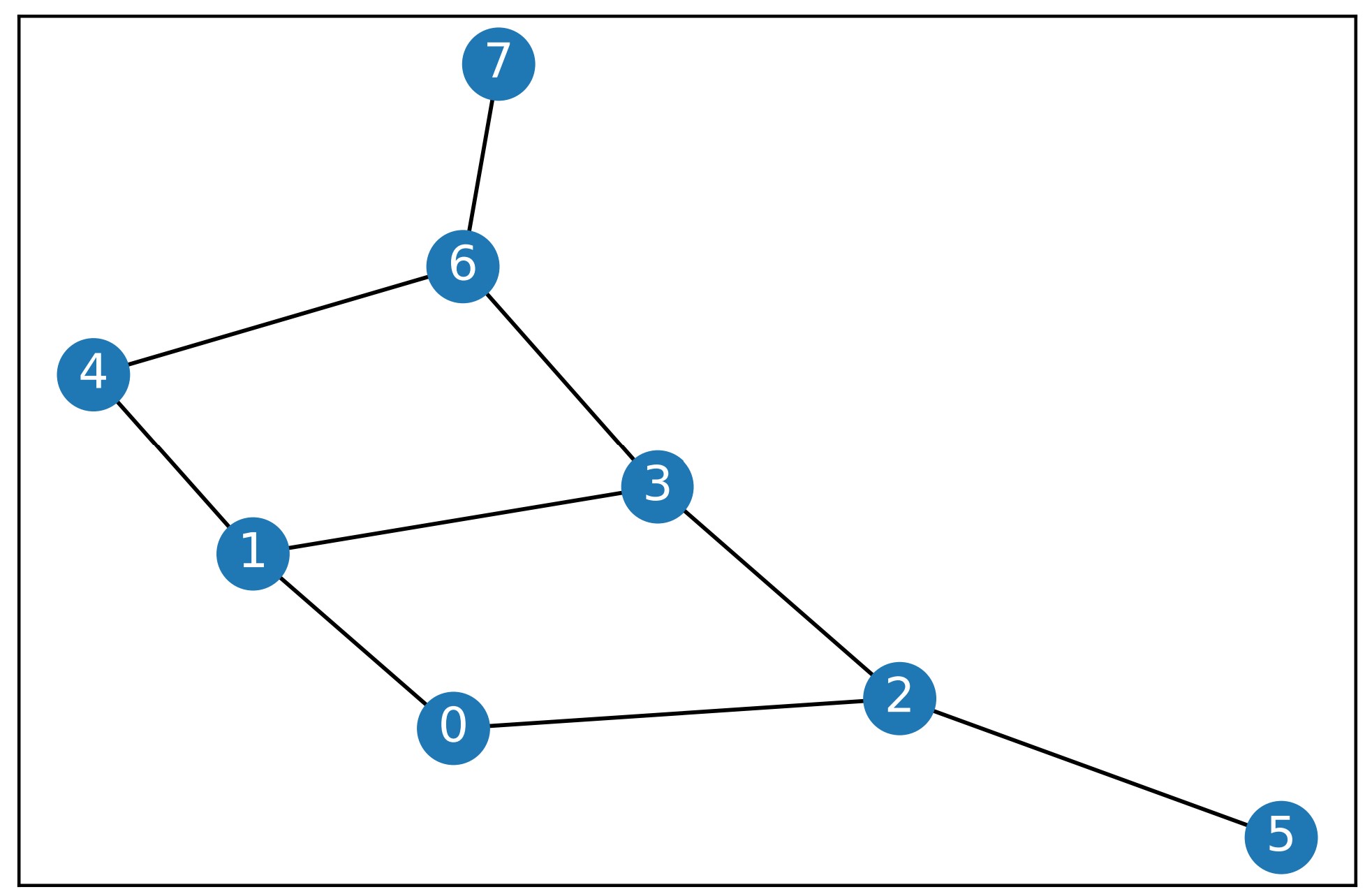} \label{fig:example_source}} 
    \subfigure[Target]{
    \includegraphics[width=0.18\textwidth]{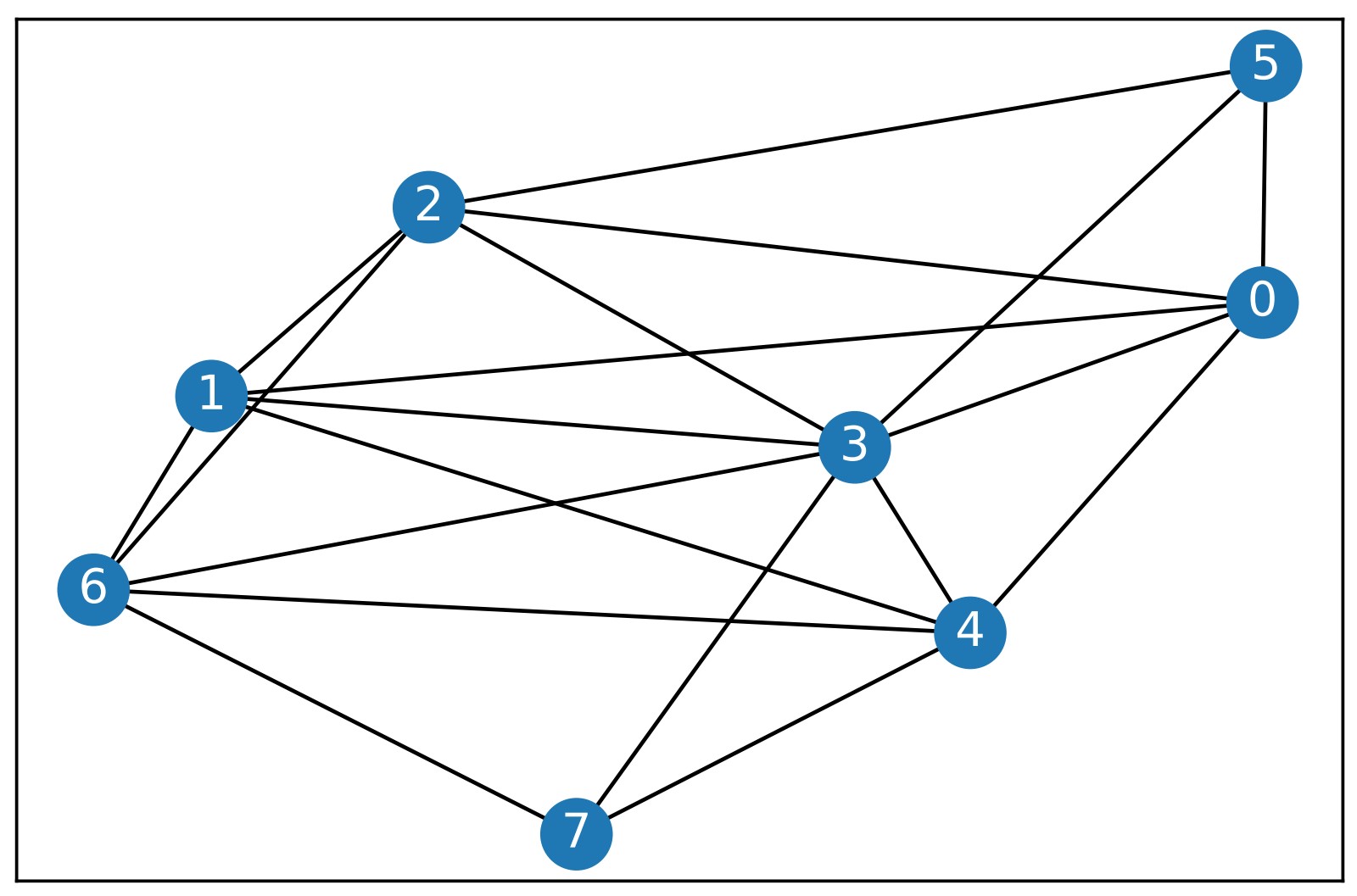}\label{fig:example_target}}
    \vskip -1em
    \subfigure[Translated by {\method}]{
    \includegraphics[width=0.18\textwidth]{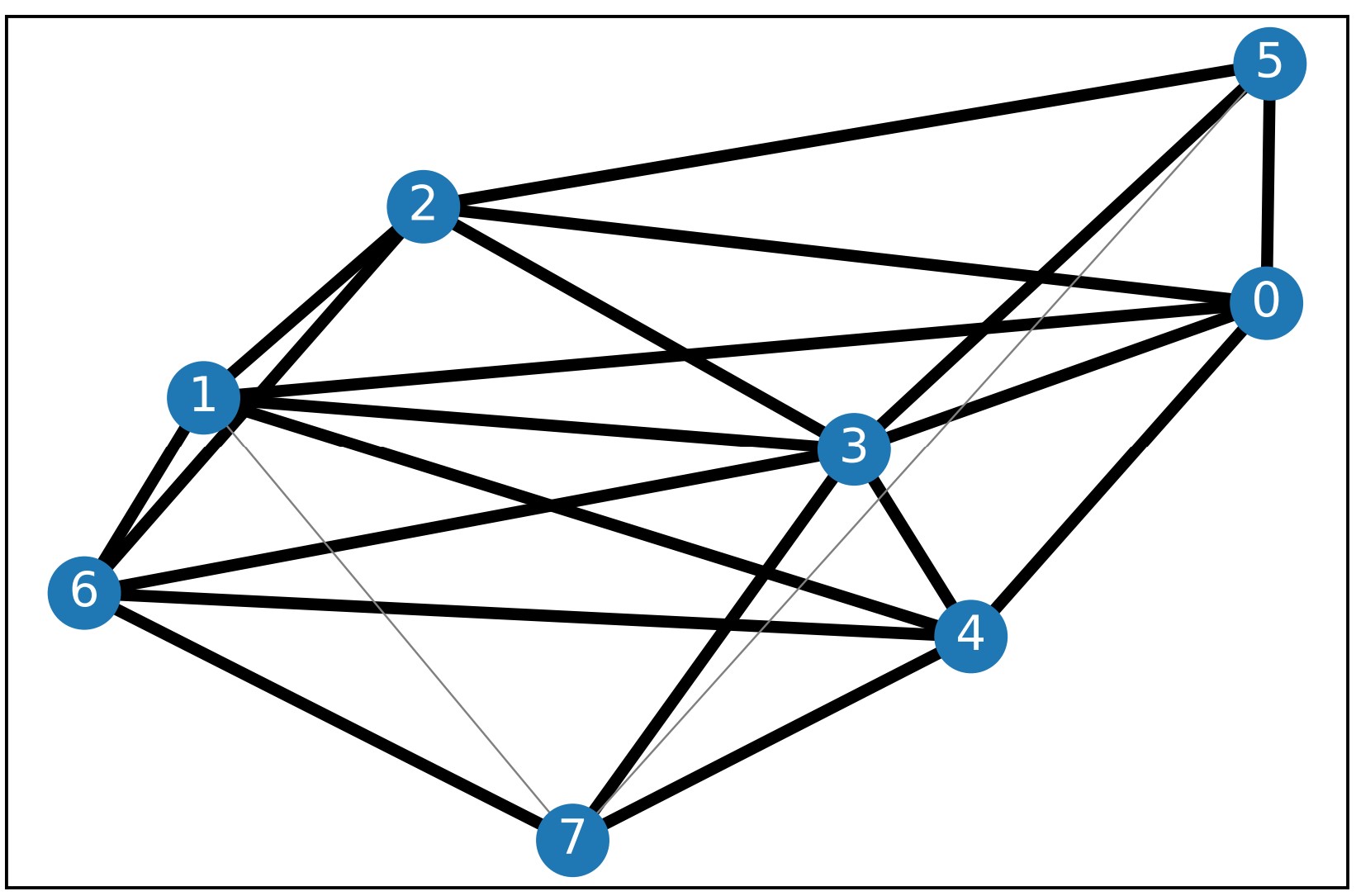}\label{fig:example_segran}}
    \subfigure[Translated by ENC-DGT-Enhanced]{
    \includegraphics[width=0.18\textwidth]{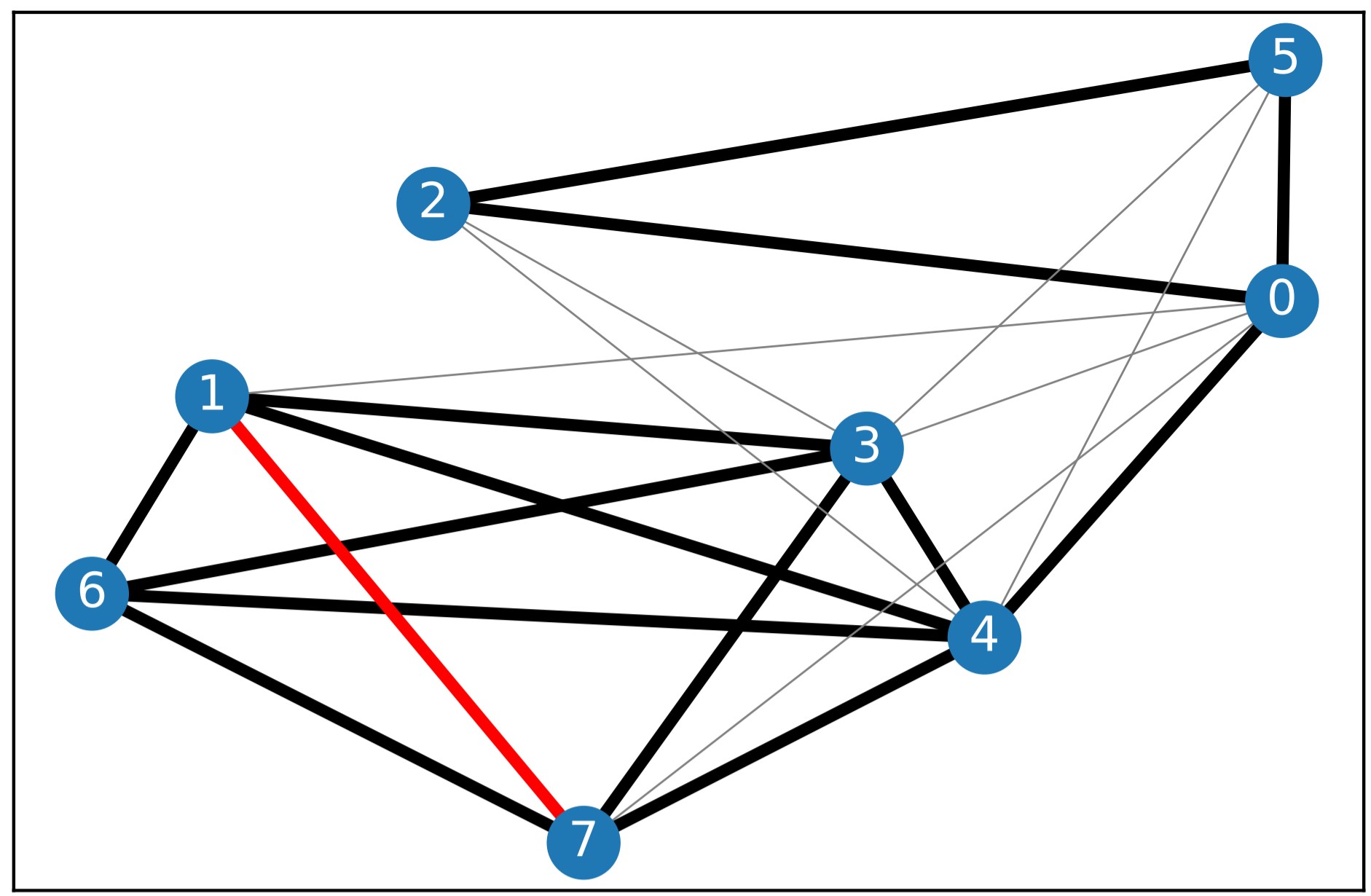}\label{fig:example_NEC}}
    \vskip -1.5em
    \caption{Obtained Case study examples.} \label{fig:case_study}
  \setlength{\abovecaptionskip}{0cm}
\end{figure}

In Figure~\ref{fig:case_study}, we provide an example of BA dataset to show and compare the captured translation patterns of {\method} and NEC-DGT-Enhanced. Figure~\ref{fig:example_source} is the source graph, and Figure~\ref{fig:example_target} is the target graph. Figure~\ref{fig:example_segran} is the result translated by {\method}, and Figure~\ref{fig:example_NEC} is translated by NEC-DGT-Enhanced. For edges in Figure~\ref{fig:example_segran} and Figure~\ref{fig:example_NEC}, to show the prediction results clearly, we split them into three groups. When the predicted existence probability is above $0.2$, we draw it in black if it is true otherwise in red. When the probability is between $0.05$ and $0.2$, we draw it in grey. Edges with smaller probability are filtered out.

It can be seen that {\method} has a high translation quality, and made no erroneous predictions. In the target graph, node $3$ is the most popular node. The same pattern can be found in the translated result by {\method}, where node $3$ has a high probability of linking to most other nodes. As to the distant node pairs, like node $5$ and $7$, or node $1$ and $7$, which have no links in the target graph, {\method} assigns a relatively low probability to their existence, below $0.2$.

The performance of NEC-DGT-Enhanced, on the other hand, is not that satisfactory. It mistakenly split the nodes into two groups, $\{0,2,5\}$ and others, and assigns large weight to edges inside them but little weight to edges between them. Node $5$ is connected only to node $2$ in the source graph, therefore NEC-DGT-Enhanced tries to push the embedding of it as well as its neighbors distant from other nodes, which could be the reason resulting in this phenomena. This example shows that {\method} is better in capturing the graph distributions in the target domain and learning the translation patterns.

\section{Conclusion and Future Work} \label{sec:conclusion}
Graph-to-graph translation has many applications. However, for many domains, obtaining large-scale data is expensive. Thus, in this paper, we investigate a new problem of semi-supervised graph-to-graph translation. We propose a new framework {\method}, which is composed of a encoder to learn graph representation, a decoder to reconstruct and a translator module to translate graph in the latent space. {\method} adopts dual embedding to bridge the semantic gap between the source and target domain. Experimental results on synthetic and real-world datasets demonstrate the effectiveness of the proposed framework for semi-supervised graph translation. Further experiments are conducted to understand $\method$ and its hyper-parameter sensitivity.

There are several interesting directions need further investigation. First in this paper, we mainly focus on translating one graph in the source domain to another graph in the target domain. In real-world, there are many situations which require the translation from one domain to many other domains. For example, in molecular translation task~\cite{Jin2018JunctionTV,Jin2019LearningMG}, we could make different requirements on the properties of translated compound, and each requirement would form a domain. Therefore, we plan to extend {\method} to one-to-many graph translation. 
Second, the graph translation has many real-world applications. In this paper, we conduct experiments on citation network and traffic network. We would like to extend our framework for more application domains such as the graph translation in brain network and medical domains~\cite{tewarie2015minimum,bassett2017network}.

\section{Acknowledgement}
This project was partially supported by NSF projects IIS-1707548, IIS-1909702, IIS-1955851, CBET-1638320 and Global Research Outreach program of Samsung Advanced Institute of Technology \#225003.

\bibliographystyle{ACM-Reference-Format}
\bibliography{dual_embedding}

\end{document}